\documentclass[letterpaper, 10 pt, conference]{ieeeconf}  

\IEEEoverridecommandlockouts                              

\usepackage{graphicx} 
\graphicspath{{figures/}}
\usepackage{amsmath} 
\usepackage{amssymb}  
\usepackage{xcolor}
\usepackage{makecell}
\usepackage[hidelinks]{hyperref} 
\usepackage{url}

\usepackage{amsthm}

\newtheoremstyle{remarkstyle}
  {\topsep}      
  {\topsep}      
  {\normalfont}  
  {}             
  {\bfseries}    
  {.}            
  { }            
  {}             
\theoremstyle{remarkstyle}
\newtheorem{remark}{Remark}

\title{\LARGE \bf
Adaptive Tuning of Parameterized Traffic Controllers via Multi-Agent Reinforcement Learning
}

\author{Giray \"{O}n\"{u}r, Azita Dabiri and Bart De Schutter
\thanks{This research has received funding from the European Research Council (ERC) under the European Union's Horizon 2020 research and innovation programme (Grant agreement No. 101018826 - ERC Advanced Grant CLariNet).}
\thanks{The authors are with the Delft Center for Systems and Control,  Delft University of Technology, Delft, The Netherlands \{\tt\small g.oenuer, a.dabiri, b.deschutter\}@tudelft.nl.}%
}

\begin{document}

\maketitle
\thispagestyle{empty}
\pagestyle{empty}

\begin{abstract}

Effective traffic control is essential for mitigating congestion in transportation networks. Conventional traffic management strategies, including route guidance and ramp metering, often rely on state feedback controllers, which are used for their simplicity and reactivity; however, they lack the adaptability required to cope with complex and time-varying traffic dynamics. This paper proposes a multi-agent reinforcement learning (RL) framework in which each agent adaptively tunes the parameters of a state feedback traffic controller, combining the reactivity of state feedback controllers with the adaptability of RL. By tuning parameters at a lower frequency rather than directly determining control inputs at a high frequency, the RL agents achieve improved training efficiency while maintaining adaptability to varying traffic conditions. The multi-agent structure further enhances system robustness, as local controllers can operate independently in the event of partial failures. The proposed framework is evaluated on a simulated multi-class transportation network under varying traffic conditions. Results show that the proposed multi-agent framework outperforms the no-control and fixed-parameter state feedback control cases, while performing on par with the single-agent RL-based adaptive state feedback control, but with much greater resilience to disturbances.

\end{abstract}

\section{Introduction}

The continuous growth in traffic demand is placing increasing pressure on existing transportation infrastructure, including both urban and freeway networks. When traffic demand surpasses road capacity, congestion arises, leading to substantial environmental, economic, and societal consequences. Passengers experience longer and less predictable commutes and are exposed to higher levels of vehicle emissions. To mitigate these issues, a variety of traffic control strategies have been developed and some have been successfully deployed in real-world settings \cite{de2017traffic}. For instance, traffic signal control at urban intersections has been shown to reduce the total time spent (TTS) by all vehicles, while ramp metering and route guidance are commonly applied on freeways to manage vehicle flow efficiently \cite{siri2021freeway}.

Traditional approaches, such as fixed-time traffic signal control, determine signal cycles offline to minimize delays under expected traffic conditions. These methods are straightforward to implement and maintain, but they lack the ability to respond to fluctuations in traffic demand or to unforeseen disturbances.


While traditional traffic control methods provide a baseline level of efficiency, their inability to adapt to real-time traffic fluctuations has motivated the development of traffic-responsive strategies. Parameterized controllers, which adjust control inputs such as ramp metering rates or vehicle splitting rates based on observed traffic states, represent a prominent class of such strategies. A well-known example is the ALINEA ramp metering controller \cite{papageorgiou1991alinea}, a state feedback method designed to maintain downstream freeway density at a desired level by continuously adjusting the metering rate using real-time measurements. Over the years, several enhancements have been proposed to extend its capabilities. PI-ALINEA \cite{wang2014local} manages bottlenecks located further downstream, while FF-ALINEA \cite{frejo2018feed} incorporates predictions of congestion evolution. Similar concepts have been applied to route guidance \cite{wang2001freeway}, where parameterized controllers regulate vehicle splitting rates based on traffic states such as queue lengths, vehicle densities, and waiting times.

Although these state feedback controllers are effective in reacting to traffic state changes, their performance critically depends on well-tuned parameters. Furthermore, local controllers generally operate independently, limiting their ability to coordinate across a transportation network.

Optimization-based approaches offer an alternative by computing control inputs through online optimization of predicted future traffic states, enabling coordination among multiple controllers \cite{van2018efficient,jeschke2023grammatical}. However, these methods often impose a significant computational burden and remain sensitive to model inaccuracies and external disturbances. A few studies have used RL to compensate for these limitations in transportation network control \cite{sun2024novel,airaldi2025reinforcement}, yet these approaches rely on centralized architectures, making them vulnerable to partial failures.

To overcome these limitations, this paper proposes a multi-agent RL-based adaptive parameterized traffic control framework. In this framework, RL agents are trained to adjust the parameters of state feedback controllers rather than directly computing control inputs. This design preserves the simplicity and reactivity of state feedback controllers while enabling them to adapt dynamically to changing traffic conditions. The multi-agent architecture enables coordination among local controllers in a decentralized way, similar to \cite{de2020independent}, and enhances system robustness, allowing other controllers to continue to operate even if one agent fails. We validate the framework on a multi-class transportation network, which better captures the complex dynamics of real-world transportation systems than a single-class representation, and test its performance under varying traffic conditions across different types of traffic controllers, namely dynamic route guidance and ramp metering controllers. The results show that the proposed framework effectively manages diverse traffic management strategies and exhibits greater resilience to disturbances compared to a centralized single-agent approach.

\section{Related Work}

This section reviews relevant prior work on RL for traffic control and on adaptive parameterized traffic control.

\subsection{Reinforcement Learning for Traffic Control}
RL has emerged as a promising approach for adaptive traffic management. By interacting with the environment and learning from feedback signals, RL agents can identify control strategies that maximize long-term performance. From vanilla Q-learning methods \cite{abdulhai2003reinforcement, li2017reinforcement} to advanced deep RL algorithms \cite{chu2019multi, wang2022integrated}, these techniques have shown notable success in both urban and freeway network control. RL methods offer the advantage of adapting to uncertainties without requiring an explicit traffic model. However, they face challenges including low sample efficiency, long training times, and limited guarantees of constraint adherence and performance, which hinder practical deployment in real-time traffic systems.

\subsection{Adaptive Parameterized Traffic Control}
Enhancing the adaptability of parameterized controllers has been a major focus in traffic control research. Approaches have ranged from optimization-based methods, such as using genetic algorithms to adjust the parameters of fuzzy controllers for ramp metering and variable speed limits in real time \cite{ghods2011adaptive}, to data-driven strategies that cluster weather and congestion patterns to select precomputed controller parameters \cite{chen2019adaptive}. While these methods can improve performance compared to offline tuning approaches, they are often unsuitable for real-time control of large-scale transportation networks due to the high computational burden of online optimization or their focus on tuning a single traffic controller without accounting for coordination among multiple controllers.

An alternative direction is to use RL for adaptive parameter tuning of state feedback controllers. Although RL faces challenges such as low sample efficiency and long training times, operating at a lower frequency to tune state feedback controller parameters rather than computing high-frequency traffic control inputs reduces the complexity of the learning problem.

Recently, \cite{sun2023adaptive} proposed an adaptive parameterized traffic control framework that combines RL with simple state feedback controllers. That approach leverages an RL agent to adjust controller parameters online, enabling adaptability to external disturbances while coordinating multiple local controllers. Numerical experiments demonstrated improved performance over conventional controllers under varying traffic conditions. However, \cite{sun2023adaptive} only considers the adaptive tuning of a single type of traffic controller, namely a ramp metering controller, and the RL agent selects its output from a set of discrete parameters defined by expert knowledge, limiting flexibility in adapting to diverse traffic conditions. Furthermore, \cite{sun2023adaptive} uses a single-agent centralized RL framework, which makes it vulnerable to partial failures and sensitive to disturbances affecting the central agent.

\section{Parameterized Traffic Controllers}

We consider a transportation network represented by the following discrete-time dynamical model:
\begin{equation}\label{eq:model}
\boldsymbol{x}(k+1) = F(\boldsymbol{x}(k),\boldsymbol{u}(k),\boldsymbol{d}(k)),
\end{equation}
where $\boldsymbol{x}(k) \in \mathbb{R}^{n_x}$ denotes the state of the transportation network, $\boldsymbol{u}(k) \in \mathbb{R}^{n_u}$ denotes the traffic control input, and $\boldsymbol{d}(k) \in \mathbb{R}^{n_d}$ represents the external disturbance influencing the traffic conditions (e.g., vehicle demand), with $n_x$, $n_u$, and $n_d$ denoting the dimensions of the state, control input, and disturbance vectors, respectively, and $k$ denotes the sampling step counter.

We assume that the network is managed by $N$ parameterized controllers, where each controller $i$ is responsible for generating and applying its own control action $u_i(k) \in \mathbb{R}$, with $u_i(k)$ being one component of the overall control input $\boldsymbol{u}(k)$. The parameterized traffic control law for controller $i$ can be expressed as:
\begin{equation*}
u_i(k) = f_i(\boldsymbol{x}_i(k),\boldsymbol{\theta}_i(k)),
\end{equation*}
where $\boldsymbol{x}_i(k) \in \mathbb{R}^{n_{x_i}}$ denotes the network state accessible to controller $i$ and $\boldsymbol{\theta}_i(k) \in \mathbb{R}^{n_{\theta_i}}$ denotes its parameters, with $n_{x_i}$ and $n_{\theta_i}$ denoting the dimensions of the accessible state and controller parameter vectors, respectively. Note that when no adaptive tuning is applied, $\boldsymbol{\theta}_i(k)$ remains constant for all $k$. When the parameters are tuned adaptively, their values may vary over time, allowing the controller to improve its performance in response to changing traffic conditions and external disturbances.

\begin{figure}[!t]
\centering
\includegraphics[width=0.425\textwidth]{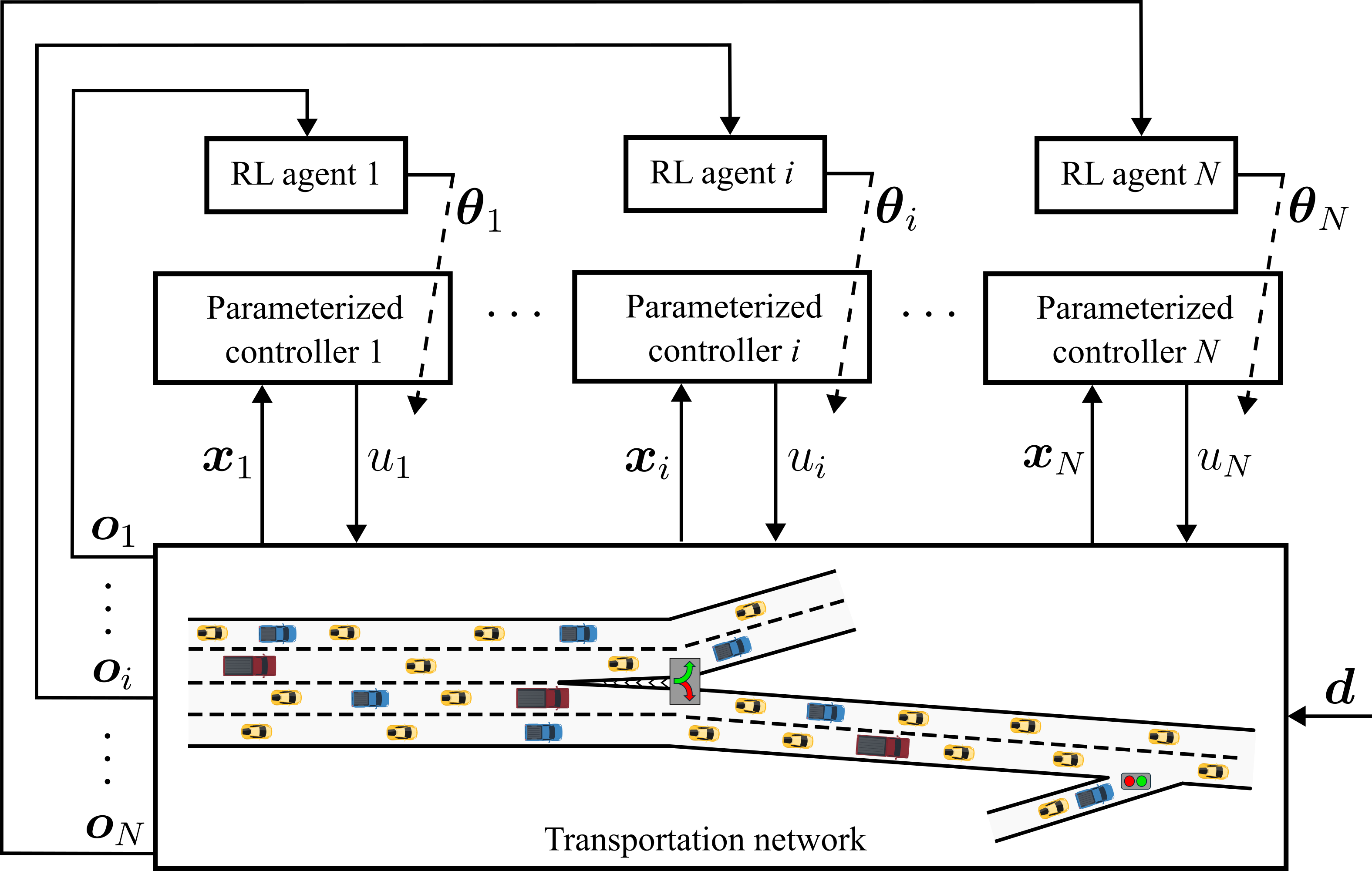}
\caption{Block diagram of the proposed multi-agent RL-based framework for adaptive tuning of parameterized controllers.} \label{fig:MARL_framework}
\end{figure}

\section{Multi-Agent RL Framework for Adaptive Parameter Tuning}

The proposed framework employs a separate RL agent for each parameterized controller to tune its parameters online, enabling adaptation to changing traffic conditions and external disturbances. The framework follows a decentralized architecture, where each RL agent uses its own local observations (see Remark~\ref{rem:2}) to determine the parameters of a traffic controller (see Figure~\ref{fig:MARL_framework}).

The framework includes two control layers. The RL agents operate at the upper layer, updating controller parameters on a slower time scale, while the state feedback controllers operate at the lower layer, computing control inputs more frequently in response to changes in the traffic states. This hierarchical structure allows the RL agents to train efficiently by adjusting parameters over longer time periods, while the parameterized state feedback controllers maintain their reactivity.

In the decentralized multi-agent RL framework, the environment consists of the RL agents, the state feedback controllers, and the controlled transportation network. Each agent interacts with the environment to learn how to adapt the parameters of its corresponding state feedback controller using a decentralized training algorithm \cite{de2020independent}. The multi-agent framework is formulated as a Markov Decision Process (MDP) with $N$ agents, represented by the tuple $\langle S, \mathcal{A}, P, \mathcal{O}, \mathcal{R}, \Gamma \rangle$, where $S$ denotes the state space of the environment, $\mathcal{A}$ is the action space including the actions of all the agents, $P: S \times \mathcal{A}^{N} \times S \to \mathbb{R}_{\geq 0}$ denotes the state transition probability kernel, which is implicitly defined by the transportation network model in \eqref{eq:model}, $\mathcal{O}$ denotes the observation space of the agents, and $\mathcal{R}$ and $\Gamma$ denote the reward and discount factor sets of all agents, respectively. Moreover, $\boldsymbol{o}_i \in \mathcal{O}$ and $\boldsymbol{a}_i \in \mathcal{A}$ denote the state observations and the actions of the $i$th agent, respectively. The reward function $r_i \in \mathcal{R}$ reflects the goal of the $i$th agent, such that $r_i: S \times \mathcal{A}^{N} \to \mathbb{R}$, while $\gamma_i \in \Gamma$ denotes its discount factor, which determines the relative weighting of future versus immediate rewards and has a value in the interval $[0,1)$. Note that the actions of the RL agents, $\boldsymbol{a}_i$, actually represent the updated parameters of their corresponding parameterized controllers.

\begin{remark}\label{rem:1}
Although the proposed framework is presented with a separate RL agent for each parameterized controller, it can also be configured so that a single RL agent adaptively tunes multiple controllers simultaneously. This allows grouping several controllers under one agent, offering a trade-off between potentially improved coordination among controllers and robustness to partial failures.
\end{remark}

\begin{remark}\label{rem:2}
Although each agent relies on its own local observations during decentralized training, all agents can use the same global reward function, which may depend on all the states of the environment and the actions of all agents, helping the agents learn coordinated behaviors toward a common goal. Even when the global reward function depends on the full system state and the actions of all agents, its computation does not require inter-agent communication, as a central unit with access to this information can evaluate the reward and provide it to each agent independently, while the training itself remains decentralized. Once training is complete, the agents are deployed in a fully decentralized manner, as illustrated in Figure~\ref{fig:MARL_framework}.
\end{remark}

\begin{figure}[!t]
\centering
\includegraphics[width=0.375\textwidth]{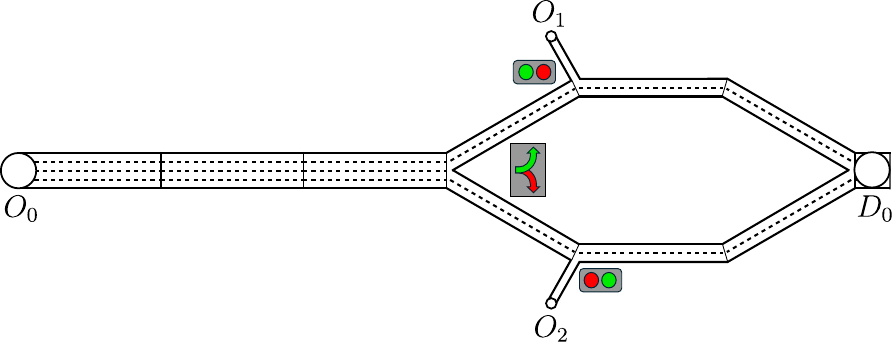}
\caption{Layout of the case study freeway network.} \label{fig:network_layout}
\end{figure}

\section{Case Study}

This section demonstrates the proposed decentralized multi-agent RL framework on a freeway traffic control case study.

\subsection{Transportation Network Setting}
We consider a freeway network with a single mainstream origin, $O_0$, that leads into two routes: a primary route with on-ramp $O_1$ and a secondary route with on-ramp $O_2$; both routes converge at the single destination, $D_0$ (see Figure~\ref{fig:network_layout}).

The network dynamics in \eqref{eq:model}, used to simulate the freeway network, are constructed based on the multi-class METANET model \cite{pasquale2016multi}, a macroscopic traffic model that balances computational efficiency and modeling accuracy. For this case study, we consider two vehicle classes within the METANET model, denoted as $\mathrm{c}_1$ and $\mathrm{c}_2$.

We consider different weather conditions to generate varying traffic patterns in the case study. To account for these conditions, certain METANET parameters are adjusted according to the weather condition, with their values provided in Table~\ref{table:params_weather}. The remaining parameters, which are independent of the weather, are listed in Table~\ref{table:params_metanet}\footnote{Parameters not explicitly specified for a given class are assumed identical for both vehicle classes.}. For further details on the multi-class METANET model and its parameters, we refer interested readers to \cite{pasquale2016multi}.

\begin{table}[!t]
\caption{METANET parameters under different weather conditions}
\begin{center}
\begin{tabular}{c c c c}
\hline

Parameter & Good Weather & Bad Weather \\
\hline
$\rho_{\mathrm{cr},\mathrm{c}_1}$ [veh/km/lane] & $40$ & $24$ \\
$\rho_{\mathrm{cr},\mathrm{c}_2}$ [veh/km/lane] & $32.65$ & $16.65$ \\
$v_{\mathrm{free},\mathrm{c}_1}$ [km/h] & $110$ & $92$ \\
$v_{\mathrm{free},\mathrm{c}_2}$ [km/h] & $86.5$ & $61.4$ \\
$\tau$ [s] & $18.0$ & $21.6$ \\
\hline

\end{tabular}
\label{table:params_weather}
\end{center}
\end{table}

\begin{table}[!t]
\setlength{\tabcolsep}{3.5pt} 
\caption{METANET parameters for the multi-class freeway network}
\begin{center}
\begin{tabular}{cccccccccc}
\hline
\noalign{\vskip 2pt}
$C_\mathrm{main}$ & $C_\mathrm{onramp}$ & $\chi$ & $\nu$ & $a^{\mathrm{c}_1}_{m}$ \\
$2000$ veh/h/lane & $2000$ veh/h/lane & $40$ veh/km/lane & $60$ km$^2$/h & $1.8$ \\
\hline
\noalign{\vskip 2pt}
$a^{\mathrm{c}_2}_{m}$ & $\delta$ & $\rho_\mathrm{max}$ & $L_m$ & $T$ \\
$2.0$ & $0.0122$ & $180$ veh/km/lane & $1000$ m & $10$ s \\
\hline
\end{tabular}
\label{table:params_metanet}
\end{center}
\end{table}

\subsection{Parameterized State Feedback Traffic Controllers}

To manage traffic flow between the primary and the secondary routes, PI-DTA \cite{wang2001freeway} is employed as a state feedback controller that dynamically adjusts vehicle flows using a PI structure to compensate for non-zero steady-state travel time difference between the two routes, defined as:
\begin{equation}\label{eq:pi-dta}
\begin{split}
u_\mathrm{dta}(k_\mathrm{dta}) = u_\mathrm{dta}(k_\mathrm{dta}-1) + K_\mathrm{P} ( \Delta T(k_\mathrm{dta}) \\ - \Delta T(k_\mathrm{dta} - 1)) + K_\mathrm{I}\Delta T(k_\mathrm{dta}),
\end{split}
\end{equation}
where $u_\mathrm{dta}$ is the proportion of flow directed toward the primary route, $\Delta T$ is the estimated travel time difference between the primary and secondary route, $K_\mathrm{P}$ and $K_\mathrm{I}$ are positive parameters, and $k_\mathrm{dta}$ is the step counter for the dynamic route guidance control. Here, $\Delta T$ is estimated as the difference between the current TTS value\footnote{For the details on the TTS calculation, we refer to \cite{pasquale2016multi}.} of vehicles on the primary and secondary routes. The tunable parameters of the controller in \eqref{eq:pi-dta} are defined as $\boldsymbol{\theta}_\mathrm{dta} = [K_\mathrm{P}, K_\mathrm{I}]^{\top}$.

To regulate on-ramp traffic and to prevent congestion, PI-ALINEA \cite{wang2014local} is applied as a state feedback method that modulates ramp flows using a PI structure that more effectively manages distant bottlenecks than standard ALINEA. The control input in PI-ALINEA is defined as:
\begin{equation}\label{eq:pi-alinea}
\begin{split}
u_\mathrm{rm}(k_\mathrm{rm}+1) = u_\mathrm{rm}(k_\mathrm{rm}) + K_\mathrm{R}(\bar{\rho}-\rho_\mathrm{b}(k_\mathrm{rm})) \\ - K_\mathrm{A}(\rho_\mathrm{b}(k_\mathrm{rm}) - \rho_\mathrm{b}(k_\mathrm{rm}-1)),
\end{split}
\end{equation}
where $u_\mathrm{rm}$ is the ramp metering rate, $\rho_\mathrm{b}$ is the lane-averaged vehicle density at a bottleneck downstream of the on-ramp, $\bar{\rho}$ is the desired downstream density parameter, $K_\mathrm{R}$ and $K_\mathrm{A}$ are positive parameters, and $k_\mathrm{rm}$ is the step counter for the ramp metering control. The tunable parameters of the controller in \eqref{eq:pi-alinea} are defined as $\boldsymbol{\theta}_\mathrm{rm} = [\bar{\rho}, K_\mathrm{R}, K_\mathrm{A}]^{\top}$.

Note that the control inputs $u_\mathrm{dta}$ and $u_\mathrm{rm}$ are constrained to the interval $[0,1]$ before being applied to the network, where $u_\mathrm{dta} = 0$ and $u_\mathrm{dta} = 1$ correspond to directing all traffic to the secondary and primary routes, respectively, while $u_\mathrm{rm} = 0$ fully blocks and $u_\mathrm{rm} = 1$ fully allows on-ramp traffic.

\subsection{RL Frameworks}

The proposed decentralized multi-agent RL-based framework and a centralized single-agent RL-based benchmark framework are implemented, where the latter serves as a baseline for comparison. In both frameworks, RL agents update the parameters of the state feedback controllers every $T_\mathrm{rl}$ seconds, which is the parameter tuning time step.

In the multi-agent framework, agent~1 tunes the parameters of the PI-DTA controller, whereas agents~2 and~3 respectively tune the parameters of the PI-ALINEA controllers on the primary and secondary routes.

The observed states of the three agents are defined as:
\begin{equation*}
\begin{split}
    \boldsymbol{o}_1 &= [\boldsymbol{d}^{\top}_1, \boldsymbol{q}^{\top}_1, \Delta T, \Delta T^{\mathrm{prev}}, u_\mathrm{dta}, w]^{\top}, \\
    \boldsymbol{o}_2 &= [\boldsymbol{d}^{\top}_2, \boldsymbol{q}^{\top}_2, \rho_\mathrm{b,1}, \rho^\mathrm{prev}_\mathrm{b,1}, u_\mathrm{rm,1}, w]^{\top}, \\
    \boldsymbol{o}_3 &= [\boldsymbol{d}^{\top}_3, \boldsymbol{q}^{\top}_3, \rho_\mathrm{b,2}, \rho^\mathrm{prev}_\mathrm{b,2}, u_\mathrm{rm,2}, w]^{\top},
\end{split}
\end{equation*}
where $\boldsymbol{d}_1$, $\boldsymbol{d}_2$, and $\boldsymbol{d}_3$ denote the current vehicle demands at origins $O_0$, $O_1$, and $O_2$, respectively, and $\boldsymbol{q}_1$, $\boldsymbol{q}_2$, and $\boldsymbol{q}_3$ represent the corresponding vehicle queue lengths, all including values for both vehicle classes. The terms $\Delta T$ and $\Delta T^{\mathrm{prev}}$ are the current and previous TTS differences between the primary and secondary routes, computed at the current and previous control steps of the PI-DTA controller. Similarly, $\rho_\mathrm{b,1}$, $\rho^\mathrm{prev}_\mathrm{b,1}$, $\rho_\mathrm{b,2}$, and $\rho^\mathrm{prev}_\mathrm{b,2}$ represent the downstream densities of the on-ramp bottlenecks evaluated at the current and previous control steps of the corresponding PI-ALINEA controllers. The variables $u_\mathrm{dta}$, $u_\mathrm{rm,1}$, and $u_\mathrm{rm,2}$ correspond to the current control inputs of the PI-DTA and PI-ALINEA controllers, while $w$ is an integer indicator representing the prevailing weather condition. The action vectors of agents 1, 2, and 3 are defined as $\boldsymbol{a}_1 = [K_\mathrm{P}, K_\mathrm{I}]^{\top}$, $\boldsymbol{a}_2 = [\bar{\rho}_1, K_\mathrm{R,1}, K_\mathrm{A,1}]^{\top}$, and $\boldsymbol{a}_3 = [\bar{\rho}_2, K_\mathrm{R,2}, K_\mathrm{A,2}]^{\top}$, respectively.

The single-agent benchmark employs a centralized adaptive tuning structure, where a single RL agent jointly learns to adjust all controller parameters based on the observed states:
\begin{equation*}
\begin{split}
    \boldsymbol{o} = [\boldsymbol{d}^{\top}_1, \boldsymbol{q}^{\top}_1, \Delta T, \Delta T^{\mathrm{prev}}, u_\mathrm{dta},\,
    \boldsymbol{d}^{\top}_2, \boldsymbol{q}^{\top}_2, \rho_\mathrm{b,1}, \\ \rho^\mathrm{prev}_\mathrm{b,1}, u_\mathrm{rm,1},\,
    \boldsymbol{d}^{\top}_3, \boldsymbol{q}^{\top}_3, \rho_\mathrm{b,2}, \rho^\mathrm{prev}_\mathrm{b,2}, u_\mathrm{rm,2}, w]^{\top},
        \end{split}
\end{equation*}
and the action vector of the single agent is defined as:
\begin{equation*}
\boldsymbol{a} = [\boldsymbol{a}^{\top}_1, \boldsymbol{a}^{\top}_2, \boldsymbol{a}^{\top}_3]^{\top}.
\end{equation*}

We implement both frameworks in MATLAB using the continuous-action actor-critic deep RL algorithm Deep Deterministic Policy Gradient (DDPG) \cite{lillicrap2015continuous}, considering its success in freeway traffic control \cite{sun2024novel,airaldi2025reinforcement}, and we train both frameworks using the hyperparameters listed in Table~\ref{table:drl_parameters}. To facilitate learning, the RL agent observations are normalized to similar orders of magnitude before being input to the actor and critic networks. The action bounds for both the multi-agent and single-agent RL frameworks are set as
    $K_\mathrm{P} \in [0,0.5],
    K_\mathrm{I} \in [0,0.1],
    K_{\mathrm{A},i} \in [0,0.1],
    K_{\mathrm{R},i} \in [0,0.05],
    \bar{\rho}_i \in [15,50]$, for  $i = 1,2.$

\begin{table}[!t]
\caption{Parameters Used for DDPG Training}
\begin{center}
\begin{tabular}{c c}
\hline
Parameter & Value \\
\hline
Number of episodes & $5000$ \\
Mini-batch size & $64$ \\
Experience replay buffer size  & $10^{4}$ \\
Discount factor & $0.99$\\
Learning rate (both actor and critic networks) & $0.001$ \\
Target network update rate & $0.01$ \\
Exploration noise standard deviation & $0.3$\\
Exploration noise decay rate & $5 \cdot 10^{-5}$ \\
\hline
\end{tabular}
\label{table:drl_parameters}
\end{center}
\end{table}

All the agents in the multi-agent and single-agent frameworks use the same global reward function:
\begin{equation*}
\begin{split}
        r(k_\mathrm{rl}) = -\sum_{k=m_\mathrm{rl} k_\mathrm{rl}}^{m_\mathrm{rl}(k_\mathrm{rl}+1)} \Big( \omega_\mathrm{TTS} J_\mathrm{TTS}(k) \\ +  \omega_\mathrm{u} \lVert \boldsymbol{u}(k) - \boldsymbol{u}(k-1) \rVert^2_2 \Big),
\end{split}
\end{equation*}
where $J_{\mathrm{TTS}}(k)$ denotes the TTS value calculated for the simulation time period $[k T, (k + 1)T)$, $\lVert \cdot \rVert_2$ denotes the Euclidean norm, $\lVert \boldsymbol{u}(k) - \boldsymbol{u}(k-1) \rVert^2_2$ is a penalty term on fluctuations between consecutive control inputs of the state feedback controllers, $m_\mathrm{rl}$ is the number of sampling steps corresponding to one adaptive tuning step, and the weights are set as $\omega_{\mathrm{TTS}} = 3.33 \cdot 10^{-4}$ and $\omega_\mathrm{u} = 2.22\cdot10^{-5}$.

\begin{figure}[!t]
\centering
\includegraphics[width=0.365\textwidth]{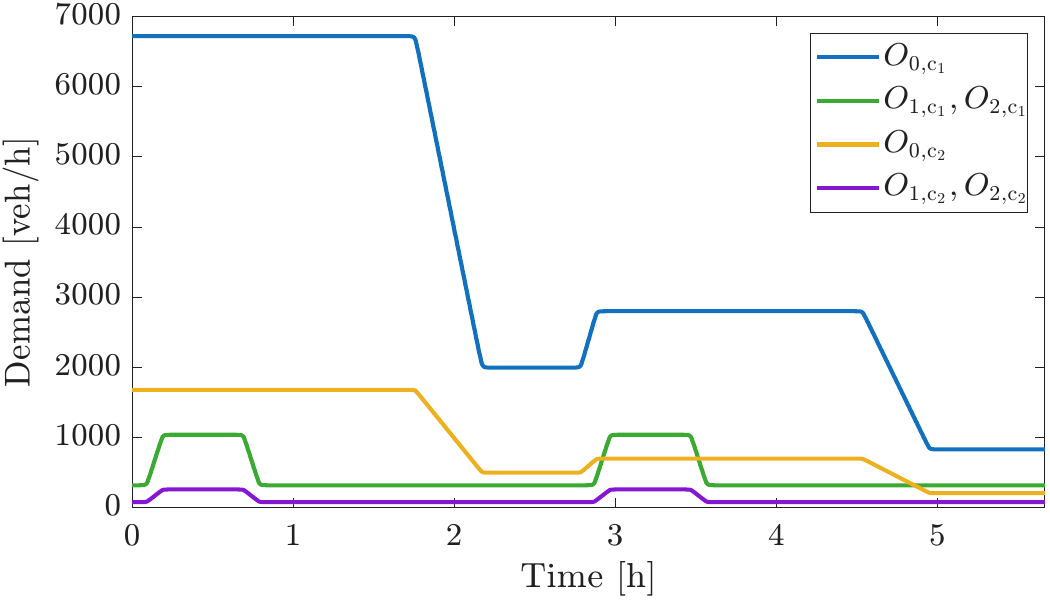}
\caption{Demand profiles for vehicle classes $\mathrm{c}_1$ and $\mathrm{c}_2$ used in the case study.} \label{fig:demand_profiles}
\end{figure}

While training the RL frameworks, we use the demand profiles given in Figure~\ref{fig:demand_profiles} for each training episode.

\subsection{Simulation Scenarios}
We simulate the case study using the multi-class METANET network model with the vehicle demands plotted in Figure~\ref{fig:demand_profiles}. We perturb the given vehicle demands by additive Gaussian noise with zero mean and origin-specific standard deviations ($\sigma_{O_{0,{\mathrm{c}_1}}}\!=\!200$, $\sigma_{O_{1,{\mathrm{c}_1}}}\!=\!\sigma_{O_{2,{\mathrm{c}_1}}}\!=\!40$, $\sigma_{O_{0,{\mathrm{c}_2}}}\!=\!50$, $\sigma_{O_{1,{\mathrm{c}_2}}}\!=\!\sigma_{O_{2,{\mathrm{c}_2}}}\!=\!10$), and smooth the noisy demand signals using a third-order low-pass Butterworth filter with a normalized cutoff frequency of $0.1$ to emulate realistic temporal dynamics.

The simulated case study lasts five and a half hours, and the weather condition shifts from good to bad at the 166th minute of the simulation.

We set the simulation time step to $T = 10\text{ s}$ (see Table~\ref{table:params_metanet}). The parameterized controllers update their control inputs via \eqref{eq:pi-dta} and \eqref{eq:pi-alinea} with sampling steps $T_\mathrm{dta} = 300\text{ s}$, $T_\mathrm{rm,1} = 60\text{ s}$, and $T_\mathrm{rm,2} = 60\text{ s}$, while the RL agents update these controllers' parameters using $T_\mathrm{rl} = 1800\text{ s}$.

In addition to the single-agent RL framework, we also use a fixed-parameter framework with hand-tuned parameters $K_\mathrm{P}=0.01$, $K_\mathrm{I}=0.005$, $K_\mathrm{A,1}=K_\mathrm{A,2}=0.1$, $K_\mathrm{R,1}=K_\mathrm{R,2}=0.005$, $\bar{\rho}_{1}=\bar{\rho}_{2}=37.5$~[veh/km/lane], and a no-control case, where the control inputs have fixed values $u_\mathrm{dta}(k)=0.5$ and $u_\mathrm{rm,1}(k)=u_\mathrm{rm,2}(k)=1$, to benchmark the performance of the proposed multi-agent RL framework.

To evaluate the robustness of the multi-agent and single-agent RL frameworks under disturbances, we conduct an additional set of experiments in which multiplicative noise is applied to the observed states associated with the PI-DTA controller, $\boldsymbol{o}_\mathrm{dta} = [\boldsymbol{d}^{\top}_1, \boldsymbol{q}^{\top}_1, \Delta T, \Delta T^{\mathrm{prev}}, u_\mathrm{dta}]^{\top}$, after their normalization and before being used as inputs to the corresponding RL agent, starting from the 30th minute of the simulation. Let $\bar{\boldsymbol{o}}_\mathrm{dta}$ denote the normalized observation vector and $\boldsymbol{\eta}$ denote the noise vector of the same dimension as $\bar{\boldsymbol{o}}_\mathrm{dta}$, with elements calculated as $\eta_i = 1 + \alpha/{100}$, where $\alpha$ is sampled from a zero-mean Gaussian distribution clipped to $[-100, 100]$, with standard deviation $\sigma \in \{0, 25, 50, 75, 100\}$ varied across experiments. The resulting noisy normalized observation is then given by
\begin{equation}\label{eq:noisy-states}
    \bar{\boldsymbol{o}}_\mathrm{dta}^{\text{noisy}} = \boldsymbol{\eta} \odot \bar{\boldsymbol{o}}_\mathrm{dta},
\end{equation}
where $\odot$ denotes elementwise multiplication.

The source code for the case study is available online\footnote{https://github.com/GirayOnur/Adaptive-Tuning-of-Parameterized-Traffic-Controllers-via-Multi-Agent-Reinforcement-Learning.}.

\begin{figure}[!t]
\centering
\includegraphics[width=0.485\textwidth]{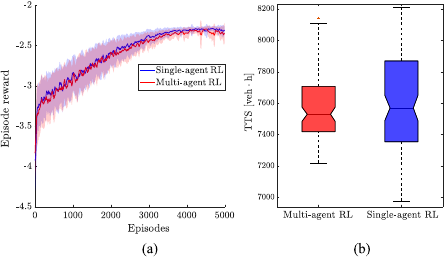}
\caption{(a) Mean and standard deviation of episode rewards across 10 training runs per setting, smoothed with a moving-average filter of size 40 to better illustrate learning progress. (b) Box plots of the TTS values for the single-agent and multi-agent settings, aggregating 100 case-study runs per setting with different random seeds, 10 runs for each of the 10 frameworks.}\label{fig:learning_curves}
\end{figure}

\begin{table}[!t]
\setlength{\tabcolsep}{4pt} 
\caption{Comparison of TTS of the proposed approach and benchmarks}
\begin{center}
\begin{tabular}{c c c c c}

\hline

  & \makecell{Multi-agent \\ RL Framework}
  & \makecell{Single-agent \\ RL Framework}
  & \makecell{Fixed \\parameters}
  & No control \\
\hline
\makecell{TTS \\ {[}veh$\cdot$h{]}} & \makecell{7544.1 \\ $\pm$ 41.1} & \makecell{7583.1 \\ $\pm$ 38.1} & \makecell{7656.7 \\ $\pm$ 42.5} & \makecell{8054.3 \\ $\pm$ 36.8} \\
\hline
\end{tabular}
\label{table:tts_all_methods}
\end{center}
\end{table}

\begin{table}[!t]
\setlength{\tabcolsep}{20pt} 
\caption{Comparison of TTS between RL frameworks under varying observation noise standard deviation $\sigma$}

\begin{center}
\begin{tabular}{c c c}
\hline

\makecell{$\sigma$}   & \makecell{Multi-agent RL}
  & \makecell{Single-agent RL} \\
\hline
0 & 7585.8 $\pm$ 235.6& 7595.0 $\pm$ 326.2 \\
25 & 7584.2 $\pm$ 228.7 & 7619.9 $\pm$ 314.3\\
50 & 7582.2 $\pm$ 227.0 & 7698.0 $\pm$ 402.0 \\
75 & 7580.8 $\pm$ 230.3 & 7747.2 $\pm$ 434.1\\
100 & 7580.8 $\pm$ 229.6 & 7789.7 $\pm$ 546.3\\
\hline
\end{tabular}
\label{table:tts_with_noise}
\end{center}
\end{table}

\subsection{Simulation Results and Discussion}

Figure~\ref{fig:learning_curves}a shows the mean and standard deviation of episode rewards for the 10 RL-based frameworks trained with different random seeds, for both the single-agent and multi-agent settings. The results show that the multi-agent frameworks have a learning convergence similar to that of the single-agent frameworks, even though they use decentralized training. Figure~\ref{fig:learning_curves}b shows the distribution of TTS values across the case-study runs. The median TTS values of the two settings are close to each other, yet the single-agent setting shows considerably higher variability compared to the multi-agent setting. A possible explanation is that the single-agent setting tunes all controller parameters jointly, which benefits from centralized decision-making and can yield better-coordinated policies than the decentralized multi-agent setting in some training runs, but also enlarges the action space and makes the training more challenging and prone to suboptimal convergence in other runs, resulting in the higher variability.

Table~\ref{table:tts_all_methods} shows the mean and standard deviation of TTS values across the case-study runs for the proposed multi-agent RL framework, the single-agent RL framework, and the fixed-parameter and no-control cases. For comparison, from the 10 trained frameworks per setting in Figure~\ref{fig:learning_curves}b, we select the one whose mean TTS is closest to the 10-framework average, to fairly represent the average performance of each RL-based method. Statistics are computed over 100 case-study runs per approach with different random seeds. The results show that adaptive RL-based methods outperform the fixed-parameter and no-control cases, while the proposed multi-agent framework performs similarly to the single-agent RL framework.

Table~\ref{table:tts_with_noise} presents the mean and standard deviation of TTS across the case-study runs for RL-based approaches under the observation noise defined in \eqref{eq:noisy-states}, using the same 10 frameworks per setting as in Figure~\ref{fig:learning_curves}b. Statistics are computed over 100 case-study runs per setting, with 10 runs per framework. The results show that TTS for the multi-agent setting remains nearly constant as noise increases, whereas both the mean and variance for the single-agent setting rise substantially. This demonstrates that the proposed decentralized multi-agent framework is resilient to disturbances, unlike the centralized single-agent RL framework.

\section{Conclusions}
We have proposed a decentralized multi-agent reinforcement learning (RL) framework that adaptively tunes the parameters of state feedback traffic controllers. This approach preserves the simplicity and reactivity of state feedback control while enhancing adaptability through RL. The decentralized structure further enhances system robustness, allowing local controllers to operate independently in the event of partial failures. Simulation results on a multi-class transportation network under varying weather conditions demonstrated that the proposed framework outperforms no-control and fixed-parameter state feedback strategies, and achieves comparable performance to centralized single-agent RL-based adaptive tuning while being more resilient to disturbances.

Future work will include extending the current decentralized framework to a distributed one, developing effective collaboration techniques among agents to achieve a cooperative framework that improves coordination while maintaining the resilience properties of the multi-agent setting.




\addtolength{\textheight}{-1cm}

\bibliographystyle{IEEEtran}
\bibliography{ref}

@article{de2020independent,
  title={Is independent learning all you need in the {S}tarcraft multi-agent challenge?},
  author={De Witt, Christian Schroeder and Gupta, Tarun and Makoviichuk, Denys and Makoviychuk, Viktor and Torr, Philip HS and Sun, Mingfei and Whiteson, Shimon},
  journal={arXiv preprint arXiv:2011.09533},
  year={2020}
}

@article{siri2021freeway,
  title={Freeway traffic control: A survey},
  author={Siri, Silvia and Pasquale, Cecilia and Sacone, Simona and Ferrara, Antonella},
  journal={Automatica},
  volume={130},
  pages={109655},
  year={2021},
  publisher={Elsevier}
}

@article{de2017traffic,
  title={Traffic management systems: A classification, review, challenges, and future perspectives},
  author={De Souza, Allan M and Brennand, Celso ARL and Yokoyama, Roberto S and Donato, Erick A and Madeira, Edmundo RM and Villas, Leandro A},
  journal={International Journal of Distributed Sensor Networks},
  volume={13},
  number={4},
  pages={1550147716683612},
  year={2017},
  publisher={Sage Publications Sage UK: London, England}
}

@article{lillicrap2015continuous,
  title={Continuous control with deep reinforcement learning},
  author={Lillicrap, Timothy P and Hunt, Jonathan J and Pritzel, Alexander and Heess, Nicolas and Erez, Tom and Tassa, Yuval and Silver, David and Wierstra, Daan},
  journal={arXiv preprint arXiv:1509.02971},
  year={2015}
}

@article{pasquale2016multi,
  title={A multi-class ramp metering and routing control scheme to reduce congestion and traffic emissions in freeway networks},
  author={Pasquale, C and Sacone, S and Siri, S and De Schutter, Bart},
  journal={IFAC-PapersOnLine},
  volume={49},
  number={3},
  pages={329--334},
  year={2016},
  publisher={Elsevier}
}

@article{sun2024novel,
  title={A novel framework combining {MPC} and deep reinforcement learning with application to freeway traffic control},
  author={Sun, Dingshan and Jamshidnejad, Anahita and De Schutter, Bart},
  journal={IEEE Transactions on Intelligent Transportation Systems},
  volume={25},
  number={7},
  pages={6756--6769},
  year={2024},
  publisher={IEEE}
}

@article{airaldi2025reinforcement,
  author={Airaldi, Filippo and Schutter, Bart De and Dabiri, Azita},
  journal={IEEE Transactions on Intelligent Transportation Systems}, 
  title={Reinforcement Learning With Model Predictive Control for Highway Ramp Metering}, 
  year={2025},
  volume={26},
  number={5},
  pages={5988-6004}

}

@article{wang2001freeway,
  title={Freeway network simulation and dynamic traffic assignment with {METANET} tools},
  author={Wang, Yibing and Messmer, Albert and Papageorgiou, Markos},
  journal={Transportation Research Record},
  volume={1776},
  number={1},
  pages={178--188},
  year={2001},
  publisher={SAGE Publications Sage CA: Los Angeles, CA}
}

@article{sun2023adaptive,
  title={Adaptive parameterized control for coordinated traffic management using reinforcement learning},
  author={Sun, Dingshan and Jamshidnejad, Anahita and De Schutter, Bart},
  journal={IFAC-PapersOnLine},
  volume={56},
  number={2},
  pages={5463--5468},
  year={2023},
  publisher={Elsevier}
}

@article{frejo2018feed,
  title={Feed-forward {ALINEA}: A ramp metering control algorithm for nearby and distant bottlenecks},
  author={Frejo, Jos{\'e} Ram{\'o}n D and De Schutter, Bart},
  journal={IEEE Transactions on Intelligent Transportation Systems},
  volume={20},
  number={7},
  pages={2448--2458},
  year={2018},
  publisher={IEEE}
}

@article{wang2014local,
  title={Local ramp metering in the presence of a distant downstream bottleneck: Theoretical analysis and simulation study},
  author={Wang, Yibing and Kosmatopoulos, Elias B and Papageorgiou, Markos and Papamichail, Ioannis},
  journal={IEEE Transactions on Intelligent Transportation Systems},
  volume={15},
  number={5},
  pages={2024--2039},
  year={2014},
  publisher={IEEE}
}

@article{papageorgiou1991alinea,
  title={{ALINEA}: A local feedback control law for on-ramp metering},
  author={Papageorgiou, Markos and Hadj-Salem, Habib and Blosseville, Jean-Marc and others},
  journal={Transportation Research Record},
  volume={1320},
  number={1},
  pages={58--67},
  year={1991}
}

@article{van2018efficient,
  title={Efficient freeway {MPC} by parameterization of {ALINEA} and a speed-limited area},
  author={van de Weg, Goof Sterk and Hegyi, Andreas and Hoogendoorn, Serge Paul and De Schutter, Bart},
  journal={IEEE Transactions on Intelligent Transportation Systems},
  volume={20},
  number={1},
  pages={16--29},
  year={2018},
  publisher={IEEE}
}

@article{jeschke2023grammatical,
  title={Grammatical-evolution-based parameterized model predictive control for urban traffic networks},
  author={Jeschke, Joost and Sun, Dingshan and Jamshidnejad, Anahita and De Schutter, Bart},
  journal={Control Engineering Practice},
  volume={132},
  pages={105431},
  year={2023},
  publisher={Elsevier}
}

@article{ghods2011adaptive,
  title={Adaptive freeway ramp metering and variable speed limit control: a genetic-fuzzy approach},
  author={Ghods, Amir Hosein and Kian, Ashkan Rahimi and Tabibi, Masoud},
  journal={IEEE Intelligent Transportation Systems Magazine},
  volume={1},
  number={1},
  pages={27--36},
  year={2011},
  publisher={IEEE}
}

@article{chen2019adaptive,
  title={Adaptive ramp metering control for urban freeway using large-scale data},
  author={Chen, Jiming and Lin, Weixin and Yang, Zidong and Li, Jianyuan and Cheng, Peng},
  journal={IEEE Transactions on Vehicular Technology},
  volume={68},
  number={10},
  pages={9507--9518},
  year={2019},
  publisher={IEEE}
}

@article{abdulhai2003reinforcement,
  title={Reinforcement learning for true adaptive traffic signal control},
  author={Abdulhai, Baher and Pringle, Rob and Karakoulas, Grigoris J},
  journal={Journal of Transportation Engineering},
  volume={129},
  number={3},
  pages={278--285},
  year={2003},
  publisher={American Society of Civil Engineers}
}

@article{li2017reinforcement,
  title={Reinforcement learning-based variable speed limit control strategy to reduce traffic congestion at freeway recurrent bottlenecks},
  author={Li, Zhibin and Liu, Pan and Xu, Chengcheng and Duan, Hui and Wang, Wei},
  journal={IEEE Transactions on Intelligent Transportation Systems},
  volume={18},
  number={11},
  pages={3204--3217},
  year={2017},
  publisher={IEEE}
}

@article{chu2019multi,
  title={Multi-agent deep reinforcement learning for large-scale traffic signal control},
  author={Chu, Tianshu and Wang, Jie and Codec{\`a}, Lara and Li, Zhaojian},
  journal={IEEE Transactions on Intelligent Transportation Systems},
  volume={21},
  number={3},
  pages={1086--1095},
  year={2019},
  publisher={IEEE}
}

@article{wang2022integrated,
  title={Integrated traffic control for freeway recurrent bottleneck based on deep reinforcement learning},
  author={Wang, Chong and Xu, Yang and Zhang, Jian and Ran, Bin},
  journal={IEEE Transactions on Intelligent Transportation Systems},
  volume={23},
  number={9},
  pages={15522--15535},
  year={2022},
  publisher={IEEE}
}

\end{document}